%% file: main.tex
\documentclass[11pt]{article}
\usepackage[margin=0.8in]{geometry}
\usepackage{amsmath,amssymb}
\usepackage{graphicx}
\usepackage{booktabs}
\usepackage{multirow}
\usepackage{array}
\usepackage{natbib}
\usepackage{hyperref}
\usepackage{xcolor}
\usepackage{caption}
\usepackage{subcaption}
\usepackage{authblk}
\usepackage{float}

\hypersetup{colorlinks=true,linkcolor=blue,citecolor=blue,urlcolor=blue}
\title{\textbf{Text-Guided Refinement of Multi-sequence Glioma Subregion Segmentation with a Vision-Language Foundation Model}}
\author[1]{Zach Eidex}
\author[1]{Yu-nong Lin}
\author[2]{Mojtaba Safari}
\author[2]{Sean Pitroda}
\author[2]{Ralph Weichselbaum}
\author[3]{Zhen Tian}
\author[1,2,3,*]{Xiaofeng Yang}

\affil[1]{Department of Biomedical Informatics, Emory University School of Medicine, Atlanta, GA, USA}
\affil[2]{Department of Radiation and Cellular Oncology, The University of Chicago, Chicago, IL, USA}
\affil[3]{Department of Radiation Oncology, Winship Cancer Institute, Emory University School of Medicine, Atlanta, GA, USA}
\affil[*]{Corresponding author: \href{mailto:xfyang@uchicago.edu}{xfyang@uchicago.edu}}

\begin{document}
\maketitle

\begin{abstract}
\textbf{Background:} Accurate delineation of glioma subregions is important for radiotherapy planning and longitudinal disease monitoring, but manual contour review and correction remain time-consuming. Task-specific segmentation models such as the self-configuring U-Net (nnU-Net) achieve strong performance but may generalize imperfectly across tumor cohorts and do not provide a native mechanism for clinician-directed text-based correction.

\textbf{Purpose:} We investigated whether a three-dimensional (3D) vision-language foundation model can be adapted for text-guided refinement of brain tumor subregion segmentation.

\textbf{Methods:} We developed a lightweight VoxTell-based refinement framework. Pretrained VoxTell generated initial subregion masks. Oracle correction prompts were built from segmentation errors and encoded target, action, location, imaging evidence, edit size, and preservation constraints. Frozen Qwen/VoxTell prompt embeddings were injected through trainable projections into VoxTell's multi-scale decoder conditioning while all other weights remained frozen. Training, validation, and testing used 901, 100, and 250 Brain Tumor Segmentation Glioma (BraTS-GLI) cases. Cross-dataset transfer was evaluated on 100 related brain tumor cases from meningioma, metastasis, pediatric tumor, and University of Pennsylvania glioblastoma (UPENN-GBM) cohorts.

\textbf{Results:} On the internal glioma test set, using post-contrast T1-weighted (T1c) input alone, correct text instructions improved mean subregion Dice similarity coefficient (DSC; enhancing tumor, edema, and necrotic/non-enhancing tumor core) from 0.774$\pm$0.158 to 0.796$\pm$0.137. Performance with correct instructions exceeded that obtained with blank prompts (0.762$\pm$0.155; Holm-adjusted \(p<0.001\), \(d_z=0.71\)) and contradictory prompts (0.770$\pm$0.163; \(p<0.001\), \(d_z=0.48\)). In cross-dataset testing, correct text instructions improved DSC from 0.527$\pm$0.287 to 0.550$\pm$0.278 and exceeded contradictory instructions (0.504$\pm$0.275; \(p<0.001\), \(d_z=0.43\)).

\textbf{Conclusion:} A 3D vision-language foundation model can be adapted for instruction-guided refinement of glioma subregion segmentations. The observed sensitivity to correct, blank, and contradictory prompts suggests that the framework performs text-dependent contour editing rather than nonspecific post-processing. These findings support further evaluation of text-guided segmentation refinement as a clinician-in-the-loop tool for targeted brain tumor contour correction.
\end{abstract}


\section{Introduction}

Accurate delineation of glioma subregions is clinically important for radiotherapy treatment planning and longitudinal disease monitoring. The enhancing tumor (ET) is primarily identified by abnormal enhancement on post-contrast T1-weighted (T1c) MRI, whereas edema and infiltrative tumor-associated abnormality are most conspicuous as hyperintensity on T2-weighted fluid-attenuated inversion recovery (T2-FLAIR) MRI. Necrotic and non-enhancing tumor core (NCR) provides additional information regarding tumor heterogeneity and disease burden. These subregions have distinct imaging appearances and different implications for treatment and diagnosis. Multiparametric magnetic resonance imaging (mp-MRI) is therefore routinely used for glioma characterization. However, manual delineation remains time-consuming and subject to intra- and interobserver variability due to small subregions and often ill-defined tumor margins, motivating the need for deep-learning approaches to brain tumor subregion segmentation.

Deep learning methods for glioma segmentation have progressed from patch-based and multi-scale convolutional neural networks to autoencoder-regularized and transformer-based 3D models, with the Brain Tumor Segmentation (BraTS) benchmark providing standardized subregion labels and common evaluation metrics \citep{menze2015brats,bakas2018brats,pereira2016bratscnn,havaei2017brain,kamnitsas2017efficient,myronenko20183d,hatamizadeh2022unetr,tang2022swinunetr,liu2023survey}. Within supervised, task-specific segmentation, nnU-Net remains a strong baseline because its self-configuring pipeline adapts preprocessing, network configuration, data augmentation, and postprocessing to each dataset \citep{isensee2021nnunet}. However, conventional task-specific networks may generalize poorly when applied to diseases, imaging protocols, or patient cohorts that differ from the training distribution. Foundation models such as Triad, VISTA3D, and MedDINOv3 aim to improve generalization by leveraging larger and more heterogeneous pretraining datasets \citep{wang2025triad,he2024vista3d,li2025meddinov3}, but automatic contours may still require substantial physician review and targeted correction. Thus, the clinically relevant task is not limited to generating a segmentation map, but also includes modifying an existing contour while preserving regions that are already acceptable.

In this study, we developed a VoxTell-based framework for text-guided refinement of glioma subregion segmentations. The framework preserves the pretrained 3D image representation and adds a lightweight trainable instruction branch that maps correction text into the multi-scale decoder-conditioning space. Reference-informed prompts were generated from observed segmentation errors and encoded the target subregion, edit direction, anatomic location, supporting imaging evidence, approximate edit size, and preservation constraints, with matched contradictory prompts used as controls. We evaluated whether the framework performed instruction-specific contour refinement rather than nonspecific post-training improvement by comparing correct, blank, and contradictory prompts, including modality-specific, data-scaling, and no-instruction experiments. We further assessed transfer in related intracranial tumor cohorts and compared text-guided VoxTell refinement with T1c-only nnU-Net, multimodal T1c plus T2-FLAIR nnU-Net, and SAT-style foundation-model baselines.
We summarize our contributions as follows:
\begin{enumerate}
    \item A lightweight adaptation of a pretrained 3D vision--language foundation model for text-guided refinement of glioma subregion segmentations.
    \item A structured prompting framework for targeted contour expansion, shrinkage, addition, and removal.
    \item An evaluation of instruction specificity using correct, blank, and contradictory prompts, supported by ablation, modality-specific, and data-scaling experiments.
    \item Internal and cross-dataset evaluation across multiple brain tumor cohorts, including comparisons with nnU-Net and foundation-model baselines.
\end{enumerate}
\section{Methods}

\subsection{Problem Formulation}

We approach text-guided contour refinement as a supervised one-step segmentation correction problem. Given a 3D MRI patch, a target subregion, and a paired reference mask \(y^k \in \{0,1\}^{H \times W \times D}\), a pretrained VoxTell model first generates an initial segmentation from the image and target prompt:
\begin{equation}
z_{0}^{k}=f_{\theta_0}(x,t_k), 
\quad 
p_{0}^{k}=\sigma(z_{0}^{k}),
\end{equation}
where \(t_k\) denotes the text prompt for the target subregion and \(\theta_0\) denotes the frozen pretrained VoxTell parameters. A correction instruction \(u^k\) is then generated from the discrepancy between the initial prediction \(p_0^k\) and the reference mask \(y^k\). The refinement model learns a function \(G_{\theta}\) that maps the image, target prompt, and correction instruction to a corrected segmentation:
\begin{equation}
\hat{z}^{k}=G_{\theta}(x,t_k,u^k),
\quad
\hat{p}^{k}=\sigma(\hat{z}^{k}),
\quad
\hat{m}^{k}=\mathbb{I}[\hat{p}^{k}>\tau].
\end{equation}
where \(\hat{z}^{k}\), \(\hat{p}^{k}\), and \(\hat{m}^{k}\) denote the refined logits, predicted probability map, and final binary segmentation mask for target subregion \(k\), respectively. The binary mask was obtained using a fixed threshold of \(\tau=0.5\). The function \(\sigma(\cdot)\) is the sigmoid activation, \(\tau\) is the segmentation threshold, and \(\mathbb{I}[\cdot]\) denotes the indicator function.
The trainable parameters \(\theta\) correspond to the added text-refinement and decoder-conditioning components, while the pretrained image representation is preserved. The model is optimized by minimizing a segmentation loss between the corrected logits \(\hat{z}^{k}\) and the reference mask \(y^k\):
\begin{equation}
\mathcal{L}_{\mathrm{total}}(\theta)
=
\mathbb{E}_{(x,k,y^k,u^k)}
\left[
\mathrm{BCEWithLogits}(\hat{z}^{k},y^k)
+
1 -
\frac{2\sum_i \sigma(\hat{z}^{k}_i)y^k_i+\epsilon}
{\sum_i \sigma(\hat{z}^{k}_i)+\sum_i y^k_i+\epsilon}
\right].
\end{equation}

Here, \(\mathrm{BCEWithLogits}\) penalizes voxel-wise classification errors, whereas the soft Dice term emphasizes spatial overlap between the corrected prediction and the reference segmentation. The correction instruction \(u^k\) is generated from the initial segmentation error, but the initial mask itself is not passed to the model as an additional image channel or learned memory input. Thus, the reported formulation tests whether language-based instructions can guide contour refinement using the MRI, the target prompt, and the correction text alone.

\subsection{Data Acquisition and Preprocessing}

The study cohort consisted of T1c and T2-weighted fluid-attenuated inversion recovery (T2-FLAIR) BraTS-GLI cases~\cite{menze2015brats,bakas2018brats} divided into 901 training cases, 100 validation cases, and 250 internal glioma test cases. Cross-dataset transfer was evaluated on 100 related intracranial tumor cases, with 25 cases randomly selected from each of four cohorts: meningioma, metastasis, pediatric tumor, and University of Pennsylvania glioblastoma (UPENN-GBM)~\cite{bakas2022upenn}; this set was used only for evaluation. The public datasets used in this study were distributed after standard preprocessing by the original dataset organizers, including skull stripping, within-subject image co-registration, and resampling to an isotropic 1 mm spatial resolution. For model input, each image volume was cropped to a \(192 \times 192 \times 192\) voxel patch. Image intensities were z-score normalized and scaled to the range \([-1,1]\). Since these public datasets were included during foundation-model pretraining, the cross-dataset experiment was interpreted as a transfer evaluation rather than a prospective external validation cohort.

\subsection{VoxTell Refinement Architecture}

The refinement model was initialized from a pretrained VoxTell checkpoint \citep{rokuss2025voxtell}. The image encoder, base target-prompt pathway, and segmentation decoder were kept frozen. Adaptation was restricted to a lightweight instruction branch that mapped natural-language correction commands into the multi-scale decoder-conditioning space.

The model used two text streams. The first stream encoded the base target prompt \(t_k\), such as ``enhancing brain tumor,'' ``brain edema,'' or ``necrotic brain tumor core,'' using the original VoxTell text-conditioning pathway. For decoder scale \(\ell\), this produced a frozen target embedding:
\begin{equation}
b_{\ell}^{k} =
P_{\ell}^{0}
\left(
T^{0}\left(\phi_{\mathrm{Qwen}}(t_k), B(x)\right)
\right),
\end{equation}
where \(\phi_{\mathrm{Qwen}}\) is the frozen Qwen text encoder, \(T^{0}\) is the pretrained VoxTell prompt transformer, and \(P_{\ell}^{0}\) maps the prompt representation to the channel dimension expected by decoder scale \(\ell\).

The second stream encodes the correction instruction \(u^k\). For an instruction \(u^k\), the frozen Qwen text encoder produced token embeddings:
\begin{equation}
E_{u} = \phi_{\mathrm{Qwen}}(u^k) \in \mathbb{R}^{L \times d_q},
\end{equation}
where \(L\) is the token length and \(d_q\) is the Qwen embedding dimension. These embeddings were projected into the VoxTell prompt dimension, processed by a copied VoxTell prompt transformer against the image bottleneck representation \(B(x)\), and projected into the same decoder-conditioning spaces:
\begin{equation}
i_{\ell}^{k} =
P_{\ell}^{\psi}
\left(
T^{\mathrm{inst}}\left(W^{\psi}E_{u}, B(x)\right)
\right).
\end{equation}
Here, \(W^{\psi}\) and \(P_{\ell}^{\psi}\) are trainable instruction-projection layers. The copied instruction prompt transformer \(T^{\mathrm{inst}}\) was initialized from VoxTell and kept frozen in the final lightweight configuration.

The structure and correction streams were combined by residual weighted addition before the frozen decoder:
\begin{equation}
\tilde{h}_{\ell}^{k} = b_{\ell}^{k} + \lambda i_{\ell}^{k},
\qquad
\lambda = \tanh(\gamma),
\end{equation}
where \(\gamma\) is a learnable scalar initialized to give an initial instruction contribution of approximately 0.1. This formulation preserves the base target prompt as the dominant semantic anchor while allowing the correction instruction to steer the decoder toward a local edit. The fused multi-scale prompt embeddings were then passed to the frozen VoxTell decoder:
\begin{equation}
\hat{z}^{k}
=
D^{0}
\left(
\{F_{\ell}(x)\}_{\ell=1}^{S},
\{\tilde{h}_{\ell}^{k}\}_{\ell=1}^{S}
\right).
\end{equation}

\begin{figure}[H]
\centering
\makebox[\textwidth][c]{\includegraphics[width=1.16\textwidth]{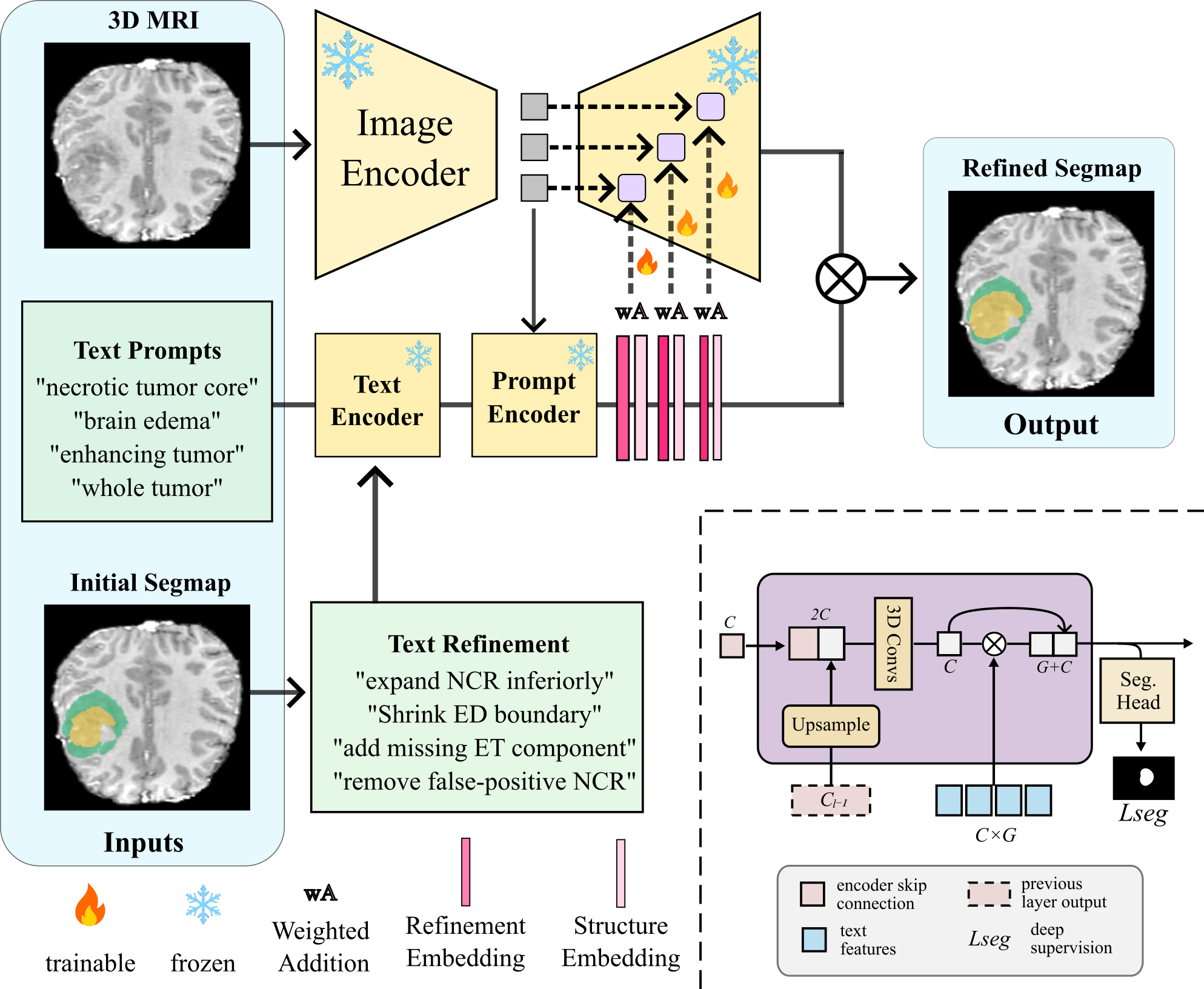}}
\caption{VoxTell text-guided refinement architecture. A frozen pretrained VoxTell pathway encodes the MRI patch and base structure prompt. A copied instruction branch encodes the correction command and maps it into the same multi-scale decoder-conditioning spaces. At each decoder scale, the structure prompt embedding and instruction embedding are fused by learned weighted addition, \(b_{\ell}^{k}+\lambda i_{\ell}^{k}\), before being passed to the frozen VoxTell decoder to produce the corrected target segmentation.}
\label{fig:architecture}
\end{figure}

\subsection{Prompt Design}

Correction prompts were generated from the disagreement between the pretrained VoxTell prediction and the reference segmentation during both model training and inference. For each target subregion \(k\), false-negative voxels represented missed target tissue and false-positive voxels represented over-segmentation. Let \(A_k\) denote the number of missed voxels and \(R_k\) denote the number of excess voxels. The requested edit was selected from four actions: expand, shrink, add, or remove. Expand and add instructions were used for missed tumor, whereas shrink and remove instructions were used for over-segmented or false-positive regions.

Each prompt was rendered as a compact clinical instruction:
\begin{equation}
u = \mathrm{Template}(k,a,r,e,s,c),
\end{equation}
where \(k\) is the target subregion, \(a\) is the requested action, \(r\) is the anatomic region, \(e\) is the supporting imaging evidence, \(s\) is the approximate edit size, and \(c\) is a preservation constraint. Target names were rendered as ``enhancing tumor'' for ET, ``edema'' for ED, and ``necrotic core'' for NCR. Location descriptors were derived from the centroid and spatial distribution of the error voxels. ET prompts emphasized rim-based locations, NCR prompts emphasized central or core-based locations, and ED prompts emphasized peripheral or outer-boundary locations.

Evidence phrases were matched to the input modality. For T1c input, ET prompts used enhancement-based language, such as ``where T1c is bright'' for expansion or addition and ``where there is no convincing T1c enhancement'' for shrinkage or removal. NCR prompts used nonenhancement or cavity-related language. Because edema is primarily evaluated on T2-FLAIR MRI, T1c-only ED prompts used more general image-support language. For T2-FLAIR-only experiments, ED prompts used FLAIR-specific language, such as ``where FLAIR is hyperintense'' or ``where there is no convincing FLAIR hyperintensity.'' Edit size was discretized as small (\(<3\) cc), medium (\(3\)--\(12\) cc), or large (\(\ge 12\) cc).

Preservation constraints were included to discourage global resegmentation. All prompts requested a local edit and preservation of other regions. Expansion and addition prompts asked the model to keep the correction attached to the current boundary and avoid distant new foci, whereas removal prompts asked the model to remove isolated false-positive regions without changing the remaining contour. Example prompts included: ``Minimal edit: Expand enhancing tumor along the superior rim where T1c is bright; keep the edit local and preserve other regions,'' and ``Minimal edit: Shrink edema at the superior outer boundary where there is no convincing FLAIR hyperintensity; keep the edit local and preserve other regions.''

For each correct prompt, a contradictory prompt was generated by reversing the requested action while preserving the target and local region when possible. Thus, expand or add instructions were paired with shrink or remove instructions, and shrink or remove instructions were paired with expand or add instructions. Each case was evaluated with correct, blank, and contradictory prompts to determine whether refinement depended on the semantic content of the instruction.

\subsection{Model Implementation and Training}

The Voxtell model was initialized from the VoxTell v1.1 checkpoint. All weights including the Qwen text embeddings, the VoxTell image encoder, the base target-prompt pathway, and the segmentation decoder were kept frozen. The final text-guided model used a lightweight projection-tuning configuration with the initial instruction scale set to 0.1. Only the copied instruction text-projection, instruction-to-decoder projection layers, and associated scalar fusion parameters were updated. The T1c-only model was trained on an Nvidia A6000 ADA GPU with 48 GB of VRAM for up to 5000 optimization steps using batch size 4, learning rate \(10^{-5}\), Adam-style weight decay \(10^{-4}\), mixed precision, and gradient clipping at 1.0. Validation was performed every 250 steps using the 100-case validation split, and the checkpoint with the lowest validation loss was used for testing. Data loading used cached \(192 \times 192 \times 192\) patches and precomputed text embeddings to dramatically reduce training time. The T2-FLAIR text-refinement model used the same training schedule and prompt-generation pipeline, with evidence phrases adapted for FLAIR input. Data-scaling and no-instruction controls used the same training pipeline, varying only the training subset size or replacing the correction instruction with a blank prompt. ET, ED, and NCR were sampled approximately equally during training.

nnU-Net comparisons used the nnU-Net v2 3D full-resolution configuration on the same data splits when available \citep{isensee2021nnunet}. T1c-only nnU-Net was used as the direct single-modality automatic segmentation baseline, and T1c plus T2-FLAIR nnU-Net was used as the multimodal automatic segmentation reference. SAT-Pro and SAT-Nano were evaluated as text-prompted foundation-model comparators rather than architecture-matched refinement models \citep{zhao2025sat}.

\subsection{Evaluation Metrics}

Segmentation performance was evaluated using the Dice similarity coefficient (DSC), normalized surface Dice (NSD), and 95th percentile Hausdorff distance (HD95). Mean subregion DSC was computed as the average of ET, ED, and NCR. Whole tumor (WT) was computed at evaluation as the union of ET, ED, and NCR and was reported separately. Aggregate values are reported as mean$\pm$standard deviation over cases.

The primary text-guidance endpoints were correct-minus-pretrained, correct-minus-blank, and correct-minus-contradictory performance differences. Correct-minus-pretrained measured improvement over the initial VoxTell prediction. Correct-minus-blank measured the added value of the correction instruction relative to the same model without correction text. Correct-minus-contradictory measured semantic specificity by comparing appropriate and inappropriate instructions for the same case and target. A successful text-guided editor was expected to satisfy:
\begin{equation}
\begin{aligned}
\mathrm{DSC}(G_{\theta}(x,t_k,u_{\mathrm{correct}}),y^k)
&>
\mathrm{DSC}(G_{\theta}(x,t_k,u_{\mathrm{blank}}),y^k),\\
\mathrm{DSC}(G_{\theta}(x,t_k,u_{\mathrm{correct}}),y^k)
&>
\mathrm{DSC}(G_{\theta}(x,t_k,u_{\mathrm{contradictory}}),y^k).
\end{aligned}
\end{equation}

Paired comparisons used two-sided paired \(t\)-tests of case-level metric differences. Multiple comparisons were controlled using Holm-Bonferroni correction within each table block and metric. Effect sizes are reported as paired Cohen's \(d_z=\bar{\Delta}/s_{\Delta}\), where \(\Delta\) is the case-level paired difference oriented so that positive values favor the proposed text-guided condition. For all methods with retained case-level outputs, displayed mean subregion metrics and statistical tests were computed from the same case-level arrays.

\section{Results}

\subsection{Text-guided Refinement}

Quantitative results for the primary VoxTell refinement experiments are summarized in Table~\ref{tab:main}. On the internal GLI test set, pretrained VoxTell achieved a mean subregion DSC of 0.774$\pm$0.158 using T1c input alone. Correct text instructions increased mean subregion DSC to 0.796$\pm$0.137, compared with 0.762$\pm$0.155 for blank prompts and 0.770$\pm$0.163 for contradictory prompts. This corresponded to DSC improvements of 0.022 over the pretrained prediction, 0.035 over blank prompts, and 0.026 over contradictory prompts. Correct text also improved NSD from 0.753$\pm$0.160 to 0.774$\pm$0.145 and reduced HD95 from 7.06$\pm$8.63 mm to 6.47$\pm$7.67 mm.

On the cross-dataset transfer set, pretrained VoxTell achieved a mean subregion DSC of 0.527$\pm$0.287. Correct text instructions increased DSC to 0.550$\pm$0.278, compared with 0.520$\pm$0.265 for blank prompts and 0.504$\pm$0.275 for contradictory prompts. The correct-minus-contradictory separation was larger cross-dataset than internally (0.047 vs. 0.026 DSC), suggesting that the instruction branch retained text-dependent behavior under dataset shift. Correct text also increased NSD from 0.532$\pm$0.249 to 0.557$\pm$0.248 and reduced HD95 from 28.42$\pm$52.17 mm to 25.97$\pm$54.64 mm. Cross-dataset WT DSC decreased from 0.732 to 0.698, indicating that some subregion-level corrections redistributed labels in a way that improved target compartments but reduced union overlap.

\begin{table}[H]
\centering
\caption{VoxTell text-guided refinement results for T1-contrast and T2-FLAIR volumes. Frozen encoder and full-finetune rows are post-trained without additional text refinement prompts or changes to the original Voxtell architecture. Holm-adjusted paired \(t\)-test p-value and paired Cohen's \(d_z\) are reported for each metrics.}
\label{tab:main}
\resizebox{\textwidth}{!}{\input{tables/main_voxtell_results.tex}}
\end{table}

At the subregion level, correct text improved internal DSC from 0.860 to 0.868 for ET, from 0.693 to 0.712 for ED, and from 0.770 to 0.808 for NCR. In the cross-dataset transfer set, correct text improved DSC from 0.725 to 0.748 for ET, from 0.353 to 0.366 for ED, and from 0.504 to 0.537 for NCR. Contradictory prompts generally degraded performance or failed to improve the target subregion, most notably for cross-dataset NCR, where DSC decreased to 0.442. Paired testing supported the observed text response. For T1c input, correct text exceeded pretrained, blank, and contradictory variants by \(\Delta\)DSC values of 0.022, 0.035, and 0.026 internally, and by 0.023, 0.030, and 0.047 cross-dataset. For T2-FLAIR input, the corresponding gains were 0.102, 0.024, and 0.085 internally, and 0.037, 0.010, and 0.049 cross-dataset. Metric-specific adjusted p-values and paired effect sizes are reported in Table~\ref{tab:main}.

\begin{figure}[H]
\centering
\makebox[\textwidth][c]{\includegraphics[width=1.20\textwidth]{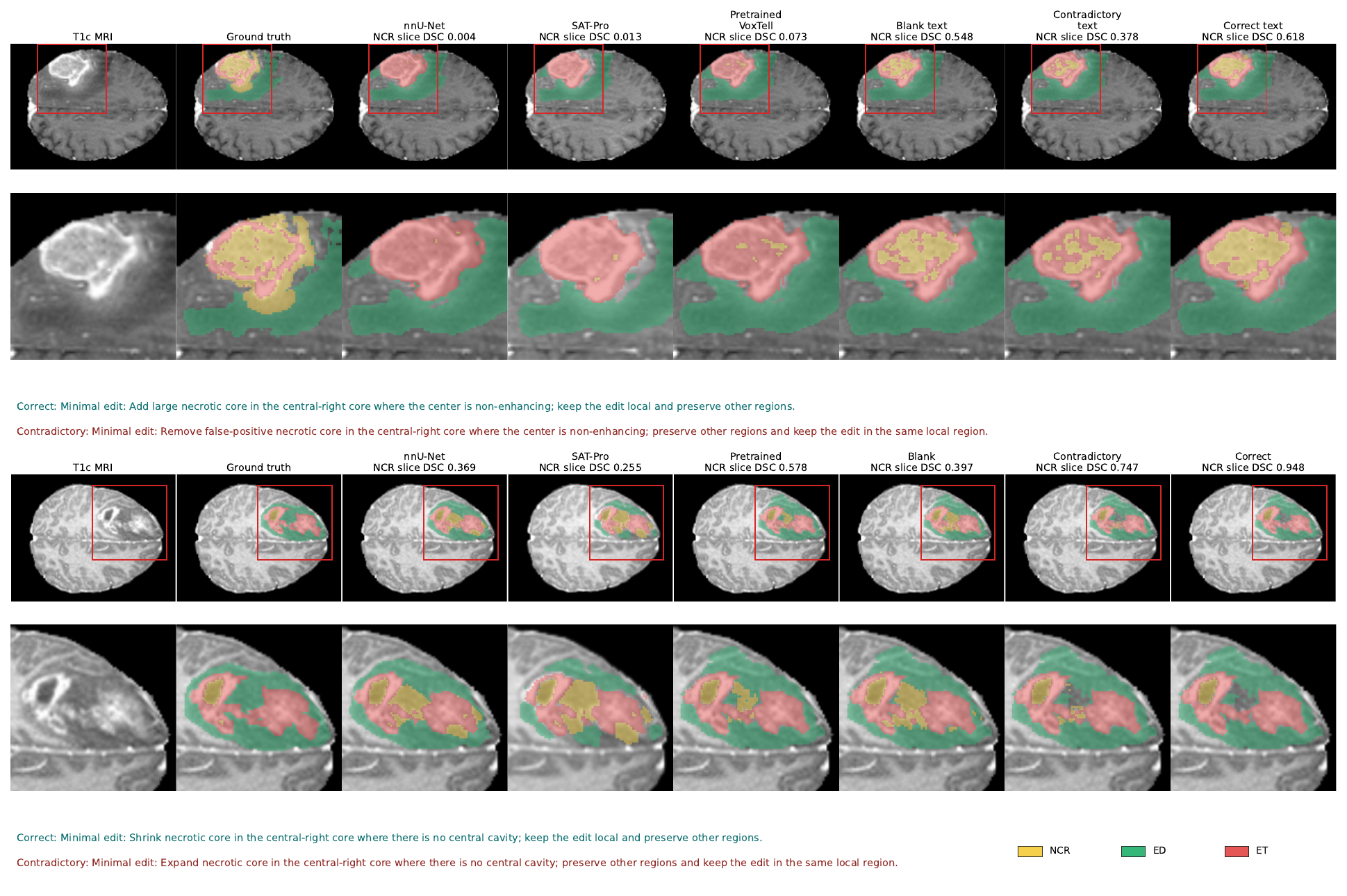}}
\caption{Internal GLI qualitative examples. Each row shows one patient and one target-specific correction instruction. All tumor compartments are shown together: NCR yellow, ED green, and ET red. Red boxes identify the common tumor zoom region used across methods. Target DSC values below the panels quantify the highlighted target response.}
\label{fig:qualinternal}
\end{figure}

\begin{figure}[H]
\centering
\makebox[\textwidth][c]{\includegraphics[width=1.20\textwidth]{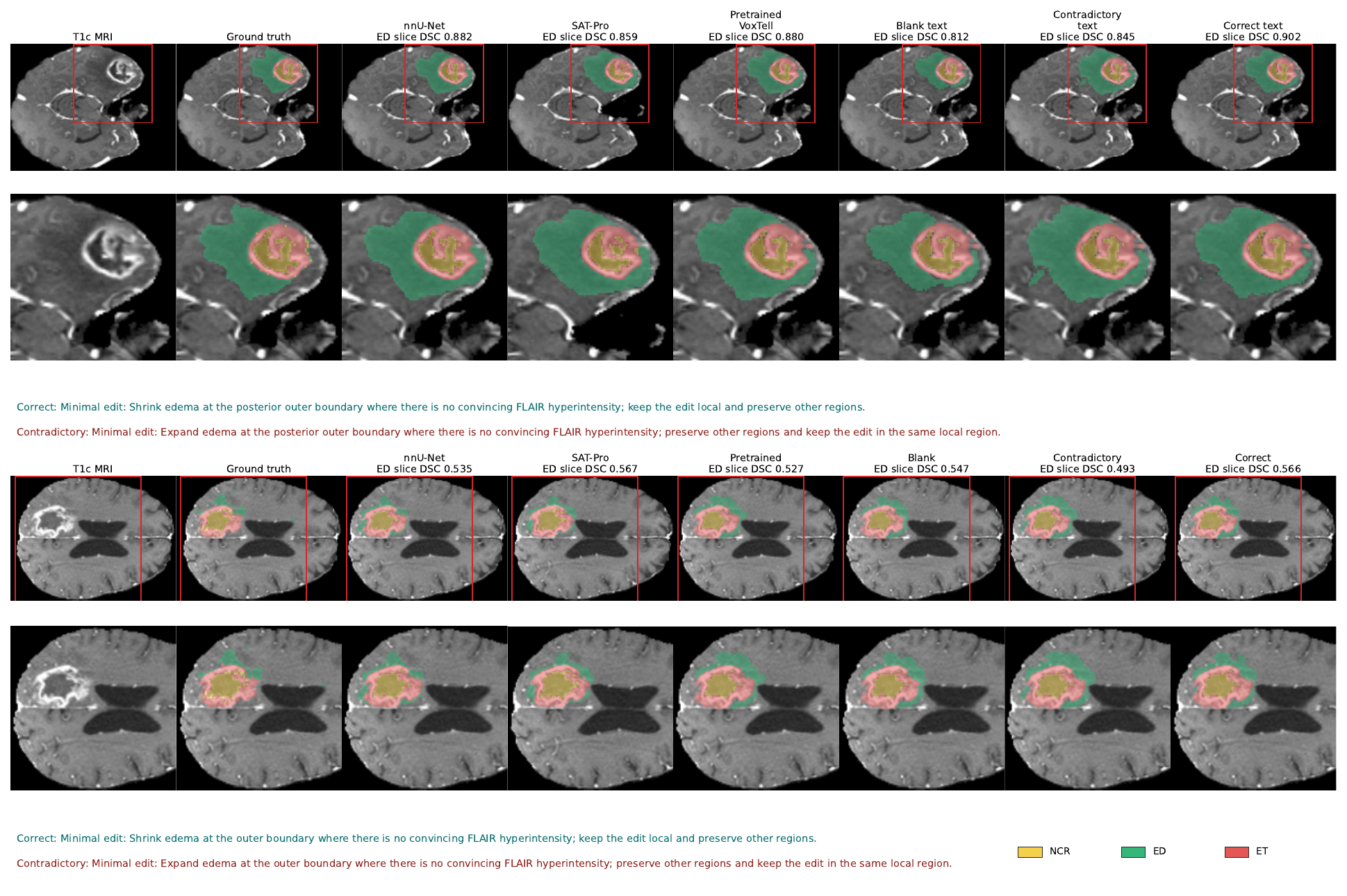}}
\caption{Cross-dataset qualitative examples. Correct text is compared with blank and contradictory text on the same image and target; SAT-Pro and T1c nnU-Net are included as foundation-model and automatic comparators. Red boxes identify the common tumor zoom region used across methods. The examples demonstrate instruction-sensitive behavior under dataset shift.}
\label{fig:qualexternal}
\end{figure}

Figures~\ref{fig:qualinternal} and~\ref{fig:qualexternal} show representative instruction-response examples. In the internal examples, correct text improved NCR DSC from 0.073 to 0.618 based on the instruction to add a large necrotic core in the central-right core region, whereas the blank prompt after post-training resulted in a DSC of 0.548 and the contradictory instruction was only 0.378. In the first cross-dataset example, instructing the model to shrink the necrotic core improved ED DSC from 0.578 to 0.948 while instead giving the contradictory text to expand the region was much lower at 0.747. These examples demonstrate that the response was not simply a generic post-training improvement. Blank and contradictory prompts often under-corrected, over-corrected, or preserved the original error.

\subsection{Comparison with Baseline Methods}

Text-guided VoxTell was evaluated as a controllable refinement model rather than a replacement for automatic segmentation. Internally, T1c correct-text VoxTell achieved a mean subregion DSC of 0.796, compared with 0.788 for T1c-only nnU-Net. The multimodal T1c+T2-FLAIR nnU-Net was the strongest internal automatic baseline, with a mean subregion DSC of 0.843. Cross-dataset, T1c correct-text VoxTell achieved a DSC of 0.550, exceeding T1c-only nnU-Net (0.485) and nearly matching multimodal nnU-Net (0.552). SAT-Pro zero-shot T1c achieved a cross-dataset mean subregion DSC of 0.566, but did not provide the text-guided correction interface evaluated for VoxTell.

T2-FLAIR-only experiments were included because edema is more conspicuous on FLAIR than on T1c. Internally, T2-FLAIR correct text improved mean subregion DSC from 0.573 to 0.675, compared with 0.650 for blank text, 0.590 for contradictory text, and 0.661 for T2-FLAIR-only nnU-Net. ED DSC improved from 0.765 to 0.803, while T2-FLAIR-only nnU-Net achieved the highest internal ED DSC at 0.812. Cross-dataset, T2-FLAIR correct text improved mean subregion DSC from 0.442 to 0.479, compared with 0.469 for blank text, 0.430 for contradictory text, and 0.415 for T2-FLAIR-only nnU-Net. As expected, T2-FLAIR remained weaker than T1c for ET and NCR, consistent with the importance of post-contrast enhancement and core appearance for these compartments.

\begin{table}[H]
\centering
\caption{Modality-specific comparison of VoxTell text refinement, no-text VoxTell controls, SAT, and nnU-Net. T1c and T2-FLAIR rows are single-modality evaluations; Best rows select the better single-modality result per subregion, except nnU-Net, which uses joint T1c+T2-FLAIR input. Holm-adjusted paired \(t\)-test p-values and paired Cohen's \(d_z\) are reported relative to VoxTell text within each block.}
\label{tab:modalityselected}
\resizebox{\textwidth}{!}{\input{tables/modality_selected_results.tex}}
\end{table}

No-instruction controls produced smaller gains than text-guided refinement. The best T1c no-instruction control, encoder frozen with decoder finetuning, improved internal mean subregion DSC from 0.774 to 0.784 and cross-dataset DSC from 0.527 to 0.534. Because correct, blank, and contradictory prompts are identical in the no-instruction setting, these controls cannot explain the correct-versus-contradictory separation observed with the instruction branch. Compared with no-text VoxTell controls, T1c correct-text VoxTell showed modest DSC differences internally and cross-dataset, whereas T2-FLAIR correct-text VoxTell showed larger gains over corresponding no-text controls. The corresponding DSC, NSD, and HD95 p-values and effect sizes are reported in Table~\ref{tab:modalityselected}.

Training-size experiments evaluated whether the instruction branch benefited from additional paired correction examples. Models were trained using 10, 25, 50, 100, and 901 training patients, with the zero-patient point corresponding to the unrefined pretrained baseline. Ten patients produced minimal improvement, whereas 25, 50, and 100 patients produced progressively larger gains. The full 901-patient model performed best, indicating that text-guided refinement benefits from exposure to a larger set of pretrained-model error patterns and corresponding correction prompts (Supplementary Figure~\ref{fig:scaling}).

\subsection{Cross-Dataset Transfer by Cohort}

Since the cross-dataset transfer set combined distinct tumor cohorts, Table~\ref{tab:externalcohort} reports performance stratified by source dataset. Each subgroup contained 25 cases. Correct T1c text exceeded contradictory T1c text in mean subregion DSC in each subgroup, indicating that the pooled transfer result was not driven by a single cohort. The strongest method varied by disease cohort: modality-selected VoxTell text achieved the highest mean DSC for the meningioma and metastasis cohorts, no-text VoxTell was highest for the pediatric tumor cohort, and nnU-Net was highest for UPENN-GBM. This pattern supports reporting both pooled transfer performance and disease-specific behavior, because cross-dataset generalization varied by tumor type and comparator.

\begin{table}[H]
\centering
\caption{Cross-dataset transfer results stratified by source cohort. Each cohort included 25 cases. Best rows select the better single-modality result per subregion, except nnU-Net, which uses joint T1c+T2-FLAIR input. Holm-adjusted paired \(t\)-test p-values and paired Cohen's \(d_z\) are reported relative to VoxTell text within each cohort.}
\label{tab:externalcohort}
\resizebox{\textwidth}{!}{\input{tables/external_cohort_results.tex}}
\end{table}

\section{Discussion}

In this study, we developed a VoxTell-based framework for text-guided refinement of glioma subregion segmentation. The primary finding was that refinement was instruction-dependent rather than a nonspecific effect of post-training. Correct prompts improved mean subregion DSC relative to pretrained VoxTell, blank prompts, and contradictory prompts in both the internal GLI and cross-dataset transfer sets, with a larger correct-versus-contradictory separation under dataset shift. This pattern is important because blank and contradictory prompts passed through the same model, but did not produce the same improvement. The qualitative examples support this interpretation: in Figure~\ref{fig:qualinternal}, correct NCR instructions produced the intended local expansion or shrinkage, whereas blank and contradictory prompts under-corrected or changed the contour in the wrong direction. Similarly, in Figure~\ref{fig:qualexternal}, correct ED prompts improved the local edema contour in transfer cases where nnU-Net, SAT-Pro, pretrained VoxTell, and contradictory text each left residual errors.

Across baselines, text-guided VoxTell was competitive but not uniformly the highest-performing method. Internally, multimodal T1c+T2-FLAIR nnU-Net remained the strongest automatic segmentation baseline, which is expected for a supervised glioma model with access to complementary enhancement and edema information. T1c text-guided VoxTell was similar to T1c-only nnU-Net internally, exceeded T1c-only nnU-Net cross-dataset, and nearly matched multimodal nnU-Net in the pooled transfer set. SAT-Pro also performed well in the cross-dataset T1c setting but does not support free-text segmentation, so no refinement prompts were injected. These results suggest that the main value of the proposed approach is not replacing a strong automatic segmenter on a well-matched cohort, but adding a controllable correction mechanism that remains useful when automatic segmentation performance degrades. The smaller gains from no-instruction controls further support this interpretation, because generic finetuning improved the masks less than text-guided refinement and could not produce correct-versus-contradictory separation.

The transfer results suggest that text refinement may be most useful when the initial segmentation is imperfect. The correct-versus-contradictory gap was larger cross-dataset than internally, and modality-selected VoxTell text achieved the highest mean DSC for the meningioma and metastasis cohorts, whereas nnU-Net was strongest for UPENN-GBM. This pattern is consistent with the expected behavior of a refinement tool: when an automatic contour is already close to the reference, there is limited room for a local correction to improve DSC, but when the model encounters a less familiar disease appearance or cohort distribution, targeted instructions can provide additional information about what should change. The pediatric cohort, where no-text VoxTell was strongest, also shows that text guidance is not universally beneficial and may depend on the quality of the initial segmentation, the error type, and whether the prompt describes a correction that is visually supported by the available modality. These cohort-level differences support evaluating text-guided refinement not only by pooled segmentation metrics, but also by disease type, initial contour quality, and whether the model follows the requested edit.

Several limitations should be acknowledged. Correction prompts were generated from reference-informed segmentation errors, which tests whether the model can learn and follow correction language but does not measure performance with clinician-authored prompts. The current framework also performs one-step refinement rather than iterative editing, and future work should evaluate state-aware or memory-based approaches that allow sequential user feedback. The cross-dataset analysis should be interpreted as transfer evaluation rather than prospective external validation because related public datasets may have been included during foundation-model pretraining, and each disease-specific subgroup included only 25 cases. Finally, nnU-Net and SAT provide important reference points but are not architecture-matched text-refinement comparators. Future studies should evaluate free-form clinician prompts, editing time, reader preference, dosimetric impact, and failure modes in realistic contour-review workflows.

\section{Conclusion}

We developed a lightweight text-guided refinement framework for glioma subregion segmentation based on the VoxTell 3D vision-language foundation model. Correct prompts improved segmentation performance over pretrained VoxTell and outperformed blank and contradictory prompts in both internal and cross-dataset evaluations, supporting instruction-specific refinement. These results may support further evaluation of text-guided segmentation refinement as a clinician-in-the-loop tool for targeted brain tumor contour correction.

\section*{Acknowledgments}

This work was supported in part by the National Institutes of Health, United States under Award Numbers R01DE033512, R01EB032680, R01CA272991, and U54CA274513.

\bibliographystyle{plainnat}
\bibliography{references}

\clearpage
\renewcommand{\thefigure}{S-\arabic{figure}}
\setcounter{figure}{0}
\section*{Supplementary Material}

\begin{figure}[H]
\centering
\includegraphics[width=0.74\textwidth]{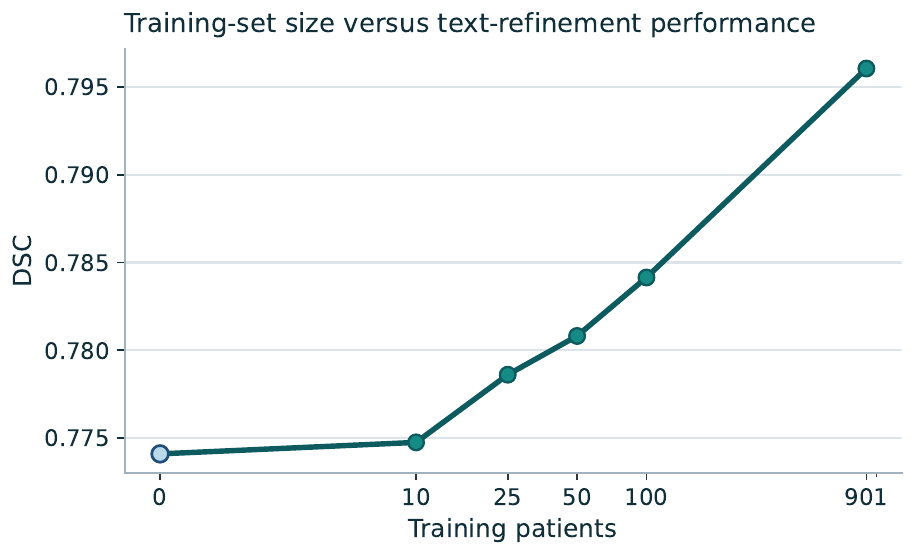}
\caption{Training-set size versus text-refinement performance. The zero-patient point shows the pretrained VoxTell segmentation before text-guided training. Subsequent points show correct-text refinement models trained with increasing numbers of patients using the same native-192 correction protocol.}
\label{fig:scaling}
\end{figure}

\end{document}

%% file: tables/main_voxtell_results.tex
\begin{tabular}{lrlrlrlrrrr}
\toprule
Variant & DSC & \(p_{\mathrm{adj}}\) (\(d_z\)) & NSD & \(p_{\mathrm{adj}}\) (\(d_z\)) & HD95 (mm) & \(p_{\mathrm{adj}}\) (\(d_z\)) & ET & ED & NCR & WT \\
\midrule
\multicolumn{11}{l}{\textbf{Internal T1c}} \\
Frozen encoder, no text & 0.784$\pm$0.156 & $<0.001$ (0.22) & 0.764$\pm$0.165 & 0.013 (0.16) & 6.71$\pm$6.65 & 0.527 (0.04) & 0.861 & 0.708 & 0.782 & 0.826 \\
Full finetune, no text & 0.777$\pm$0.166 & $<0.001$ (0.24) & 0.758$\pm$0.174 & 0.004 (0.20) & 7.60$\pm$9.58 & 0.050 (0.15) & 0.858 & 0.708 & 0.766 & 0.825 \\
Pretrained & 0.774$\pm$0.158 & $<0.001$ (0.52) & 0.753$\pm$0.160 & $<0.001$ (0.51) & 7.06$\pm$8.63 & 0.321 (0.09) & 0.860 & 0.693 & 0.770 & 0.815 \\
Blank text & 0.762$\pm$0.155 & $<0.001$ (0.71) & 0.736$\pm$0.161 & $<0.001$ (0.79) & 7.91$\pm$9.47 & 0.006 (0.21) & 0.857 & 0.666 & 0.761 & 0.805 \\
Contradictory text & 0.770$\pm$0.163 & $<0.001$ (0.48) & 0.742$\pm$0.173 & $<0.001$ (0.55) & 7.58$\pm$10.09 & 0.011 (0.19) & 0.858 & 0.690 & 0.763 & 0.811 \\
Correct text & \textbf{0.796$\pm$0.137} & - & \textbf{0.774$\pm$0.145} & - & \textbf{6.47$\pm$7.67} & - & \textbf{0.868} & \textbf{0.712} & \textbf{0.808} & \textbf{0.828} \\
\midrule
\multicolumn{11}{l}{\textbf{Cross-dataset T1c}} \\
Frozen encoder, no text & 0.534$\pm$0.266 & 0.348 (0.09) & 0.527$\pm$0.255 & 0.074 (0.18) & 35.04$\pm$62.12 & 0.180 (0.19) & 0.652 & 0.474 & 0.475 & 0.660 \\
Full finetune, no text & 0.518$\pm$0.274 & 0.157 (0.18) & 0.506$\pm$0.270 & 0.020 (0.28) & 41.05$\pm$57.23 & 0.041 (0.27) & 0.556 & \textbf{0.516} & 0.481 & 0.586 \\
Pretrained & 0.527$\pm$0.287 & 0.047 (0.25) & 0.532$\pm$0.249 & 0.023 (0.26) & 28.42$\pm$52.17 & 0.924 (0.07) & 0.725 & 0.353 & 0.504 & \textbf{0.732} \\
Blank text & 0.520$\pm$0.265 & 0.008 (0.32) & 0.510$\pm$0.259 & $<0.001$ (0.52) & \textbf{25.86$\pm$46.97} & 0.960 (-0.01) & 0.720 & 0.337 & 0.502 & 0.719 \\
Contradictory text & 0.504$\pm$0.275 & $<0.001$ (0.43) & 0.490$\pm$0.265 & $<0.001$ (0.60) & 30.26$\pm$55.17 & 0.180 (0.21) & 0.724 & 0.344 & 0.442 & 0.696 \\
Correct text & \textbf{0.550$\pm$0.278} & - & \textbf{0.557$\pm$0.248} & - & 25.97$\pm$54.64 & - & \textbf{0.748} & 0.366 & \textbf{0.537} & 0.698 \\
\midrule
\multicolumn{11}{l}{\textbf{Internal T2-FLAIR}} \\
Frozen encoder, no text & 0.646$\pm$0.147 & $<0.001$ (0.47) & 0.563$\pm$0.147 & $<0.001$ (0.27) & 10.45$\pm$10.06 & 0.020 (0.15) & 0.600 & 0.796 & 0.543 & 0.915 \\
Full finetune, no text & 0.650$\pm$0.161 & $<0.001$ (0.34) & 0.576$\pm$0.160 & 0.284 (0.07) & 11.02$\pm$10.18 & 0.004 (0.21) & 0.618 & 0.786 & 0.547 & 0.905 \\
Pretrained & 0.573$\pm$0.147 & $<0.001$ (1.39) & 0.481$\pm$0.126 & $<0.001$ (1.24) & 12.35$\pm$9.10 & $<0.001$ (0.42) & 0.614 & 0.765 & 0.339 & 0.916 \\
Blank text & 0.650$\pm$0.137 & $<0.001$ (0.51) & 0.539$\pm$0.135 & $<0.001$ (0.76) & 10.58$\pm$9.06 & 0.009 (0.18) & 0.619 & 0.783 & 0.549 & 0.914 \\
Contradictory text & 0.590$\pm$0.156 & $<0.001$ (1.01) & 0.493$\pm$0.140 & $<0.001$ (1.06) & 13.17$\pm$10.19 & $<0.001$ (0.52) & 0.602 & 0.768 & 0.399 & 0.910 \\
Correct text & \textbf{0.675$\pm$0.130} & - & \textbf{0.582$\pm$0.135} & - & \textbf{9.45$\pm$8.44} & - & \textbf{0.635} & \textbf{0.803} & \textbf{0.586} & \textbf{0.917} \\
\midrule
\multicolumn{11}{l}{\textbf{Cross-dataset T2-FLAIR}} \\
Frozen encoder, no text & 0.415$\pm$0.256 & 0.003 (0.33) & 0.385$\pm$0.240 & 0.044 (0.26) & 39.15$\pm$55.13 & 0.010 (0.32) & 0.336 & \textbf{0.558} & 0.352 & 0.664 \\
Full finetune, no text & 0.421$\pm$0.262 & 0.003 (0.34) & 0.391$\pm$0.247 & 0.044 (0.25) & 61.36$\pm$81.36 & $<0.001$ (0.46) & 0.340 & 0.540 & 0.383 & 0.628 \\
Pretrained & 0.442$\pm$0.234 & 0.001 (0.37) & 0.408$\pm$0.199 & 0.044 (0.24) & \textbf{23.96$\pm$34.33} & $>0.99$ (-0.07) & 0.487 & 0.478 & 0.360 & \textbf{0.735} \\
Blank text & 0.469$\pm$0.238 & 0.176 (0.14) & 0.416$\pm$0.226 & 0.044 (0.22) & 24.52$\pm$35.50 & $>0.99$ (-0.06) & 0.481 & 0.465 & 0.460 & 0.733 \\
Contradictory text & 0.430$\pm$0.216 & $<0.001$ (0.67) & 0.379$\pm$0.199 & $<0.001$ (0.71) & 26.43$\pm$36.42 & $>0.99$ (0.08) & 0.466 & 0.422 & 0.403 & 0.726 \\
Correct text & \textbf{0.479$\pm$0.236} & - & \textbf{0.434$\pm$0.227} & - & 25.01$\pm$36.36 & - & \textbf{0.498} & 0.460 & \textbf{0.478} & 0.732 \\
\bottomrule
\end{tabular}

%% file: tables/modality_selected_results.tex
\begin{tabular}{llrlrlrlrrrr}
\toprule
Split & Method & DSC & \(p_{\mathrm{adj}}\) (\(d_z\)) & NSD & \(p_{\mathrm{adj}}\) (\(d_z\)) & HD95 (mm) & \(p_{\mathrm{adj}}\) (\(d_z\)) & ET & ED & NCR & WT \\
\midrule
\multicolumn{12}{l}{\textbf{Internal T1c}} \\
 & nnU-Net & 0.788$\pm$0.171 & 0.082 (0.11) & \textbf{0.781$\pm$0.175} & 0.138 (-0.09) & 7.40$\pm$9.89 & 0.240 (0.12) & 0.859 & \textbf{0.713} & 0.793 & \textbf{0.833} \\
 & SAT-Pro & 0.729$\pm$0.191 & $<0.001$ (0.71) & 0.742$\pm$0.190 & $<0.001$ (0.33) & \textbf{6.37$\pm$6.04} & $>0.99$ (-0.01) & 0.805 & 0.651 & 0.731 & 0.798 \\
 & SAT-Nano & 0.712$\pm$0.201 & $<0.001$ (0.75) & 0.718$\pm$0.198 & $<0.001$ (0.49) & 7.13$\pm$7.51 & 0.552 (0.08) & 0.786 & 0.617 & 0.733 & 0.766 \\
 & VoxTell no text, frozen enc. & 0.784$\pm$0.156 & 0.001 (0.22) & 0.764$\pm$0.165 & 0.026 (0.16) & 6.71$\pm$6.65 & $>0.99$ (0.04) & 0.861 & 0.708 & 0.782 & 0.826 \\
 & VoxTell no text, full FT & 0.777$\pm$0.166 & $<0.001$ (0.24) & 0.758$\pm$0.174 & 0.005 (0.20) & 7.60$\pm$9.58 & 0.083 (0.15) & 0.858 & 0.708 & 0.766 & 0.825 \\
 & VoxTell text & \textbf{0.796$\pm$0.137} & - & 0.774$\pm$0.145 & - & 6.47$\pm$7.67 & - & \textbf{0.868} & 0.712 & \textbf{0.808} & 0.828 \\
\midrule
\multicolumn{12}{l}{\textbf{Cross-dataset T1c}} \\
 & nnU-Net & 0.485$\pm$0.327 & 0.007 (0.32) & 0.473$\pm$0.322 & $<0.001$ (0.41) & 53.64$\pm$76.49 & 0.005 (0.34) & 0.546 & 0.431 & 0.478 & 0.535 \\
 & SAT-Pro & 0.537$\pm$0.274 & 0.696 (0.06) & \textbf{0.560$\pm$0.282} & 0.890 (-0.01) & \textbf{15.17$\pm$21.15} & 0.089 (-0.22) & 0.603 & \textbf{0.524} & 0.484 & 0.655 \\
 & SAT-Nano & 0.483$\pm$0.305 & 0.005 (0.34) & 0.498$\pm$0.305 & 0.021 (0.29) & 18.54$\pm$20.45 & 0.160 (-0.14) & 0.552 & 0.387 & 0.509 & 0.578 \\
 & VoxTell no text, frozen enc. & 0.534$\pm$0.266 & 0.696 (0.09) & 0.527$\pm$0.255 & 0.148 (0.18) & 35.04$\pm$62.12 & 0.109 (0.19) & 0.652 & 0.474 & 0.475 & 0.660 \\
 & VoxTell no text, full FT & 0.518$\pm$0.274 & 0.235 (0.18) & 0.506$\pm$0.270 & 0.021 (0.28) & 41.05$\pm$57.23 & 0.032 (0.27) & 0.556 & 0.516 & 0.481 & 0.586 \\
 & VoxTell text & \textbf{0.550$\pm$0.278} & - & 0.557$\pm$0.248 & - & 25.97$\pm$54.64 & - & \textbf{0.748} & 0.366 & \textbf{0.537} & \textbf{0.698} \\
\midrule
\multicolumn{12}{l}{\textbf{Internal T2-FLAIR}} \\
 & nnU-Net & 0.661$\pm$0.161 & 0.016 (0.15) & \textbf{0.616$\pm$0.163} & $<0.001$ (-0.36) & 10.74$\pm$10.74 & 0.066 (0.15) & 0.630 & \textbf{0.812} & 0.543 & \textbf{0.927} \\
 & SAT-Pro & 0.571$\pm$0.170 & $<0.001$ (0.92) & 0.544$\pm$0.151 & $<0.001$ (0.33) & 9.50$\pm$8.04 & 0.921 (0.01) & 0.535 & 0.722 & 0.457 & 0.894 \\
 & SAT-Nano & 0.538$\pm$0.174 & $<0.001$ (1.11) & 0.504$\pm$0.149 & $<0.001$ (0.61) & 11.07$\pm$10.63 & 0.066 (0.14) & 0.494 & 0.689 & 0.430 & 0.879 \\
 & VoxTell no text, frozen enc. & 0.646$\pm$0.147 & $<0.001$ (0.47) & 0.563$\pm$0.147 & $<0.001$ (0.27) & 10.45$\pm$10.06 & 0.066 (0.15) & 0.600 & 0.796 & 0.543 & 0.915 \\
 & VoxTell no text, full FT & 0.650$\pm$0.161 & $<0.001$ (0.34) & 0.576$\pm$0.160 & 0.284 (0.07) & 11.02$\pm$10.18 & 0.007 (0.21) & 0.618 & 0.786 & 0.547 & 0.905 \\
 & VoxTell text & \textbf{0.675$\pm$0.130} & - & 0.582$\pm$0.135 & - & \textbf{9.45$\pm$8.44} & - & \textbf{0.635} & 0.803 & \textbf{0.586} & 0.917 \\
\midrule
\multicolumn{12}{l}{\textbf{Cross-dataset T2-FLAIR}} \\
 & nnU-Net & 0.415$\pm$0.280 & 0.004 (0.35) & 0.395$\pm$0.269 & 0.132 (0.20) & 53.38$\pm$71.31 & 0.003 (0.36) & 0.321 & 0.546 & 0.380 & 0.540 \\
 & SAT-Pro & 0.467$\pm$0.223 & 0.469 (0.07) & \textbf{0.471$\pm$0.213} & 0.132 (-0.20) & \textbf{15.20$\pm$21.31} & 0.014 (-0.28) & 0.443 & \textbf{0.616} & 0.341 & \textbf{0.761} \\
 & SAT-Nano & 0.425$\pm$0.240 & 0.017 (0.27) & 0.425$\pm$0.228 & 0.665 (0.04) & 20.62$\pm$21.71 & 0.189 (-0.13) & 0.385 & 0.527 & 0.364 & 0.676 \\
 & VoxTell no text, frozen enc. & 0.415$\pm$0.256 & 0.004 (0.33) & 0.385$\pm$0.240 & 0.055 (0.26) & 39.15$\pm$55.13 & 0.008 (0.32) & 0.336 & 0.558 & 0.352 & 0.664 \\
 & VoxTell no text, full FT & 0.421$\pm$0.262 & 0.004 (0.34) & 0.391$\pm$0.247 & 0.058 (0.25) & 61.36$\pm$81.36 & $<0.001$ (0.46) & 0.340 & 0.540 & 0.383 & 0.628 \\
 & VoxTell text & \textbf{0.479$\pm$0.236} & - & 0.434$\pm$0.227 & - & 25.01$\pm$36.36 & - & \textbf{0.498} & 0.460 & \textbf{0.478} & 0.732 \\
\midrule
\multicolumn{12}{l}{\textbf{Internal Best}} \\
 & nnU-Net & \textbf{0.843$\pm$0.154} & 0.006 (-0.17) & \textbf{0.858$\pm$0.162} & $<0.001$ (-0.39) & 5.42$\pm$8.20 & 0.967 (0.04) & \textbf{0.873} & \textbf{0.863} & 0.793 & \textbf{0.930} \\
 & SAT-Pro & 0.768$\pm$0.162 & $<0.001$ (0.88) & 0.790$\pm$0.165 & $<0.001$ (0.47) & 5.62$\pm$5.49 & 0.322 (0.10) & 0.811 & 0.746 & 0.746 & 0.902 \\
 & SAT-Nano & 0.754$\pm$0.166 & $<0.001$ (0.90) & 0.770$\pm$0.166 & $<0.001$ (0.63) & 5.80$\pm$5.36 & 0.279 (0.12) & 0.797 & 0.720 & 0.744 & 0.893 \\
 & VoxTell no text, frozen enc. & 0.820$\pm$0.137 & 0.004 (0.20) & 0.817$\pm$0.147 & 0.005 (0.18) & 5.35$\pm$5.21 & 0.967 (0.04) & 0.863 & 0.801 & 0.796 & 0.916 \\
 & VoxTell no text, full FT & 0.814$\pm$0.141 & $<0.001$ (0.26) & 0.810$\pm$0.152 & $<0.001$ (0.26) & 6.11$\pm$7.41 & 0.059 (0.16) & 0.859 & 0.798 & 0.786 & 0.906 \\
 & VoxTell text & 0.830$\pm$0.122 & - & 0.827$\pm$0.132 & - & \textbf{5.16$\pm$5.80} & - & 0.870 & 0.807 & \textbf{0.813} & 0.918 \\
\midrule
\multicolumn{12}{l}{\textbf{Cross-dataset Best}} \\
 & nnU-Net & 0.552$\pm$0.328 & 0.013 (0.31) & 0.562$\pm$0.334 & 0.095 (0.23) & 39.72$\pm$63.40 & 0.006 (0.32) & 0.556 & 0.571 & 0.531 & 0.624 \\
 & SAT-Pro & 0.610$\pm$0.254 & $>0.99$ (0.02) & \textbf{0.635$\pm$0.270} & 0.777 (-0.11) & \textbf{12.45$\pm$19.10} & 0.088 (-0.20) & 0.604 & \textbf{0.673} & 0.553 & 0.787 \\
 & SAT-Nano & 0.566$\pm$0.291 & 0.038 (0.26) & 0.593$\pm$0.296 & 0.777 (0.11) & 15.96$\pm$19.20 & 0.385 (-0.09) & 0.574 & 0.556 & 0.567 & 0.723 \\
 & VoxTell no text, frozen enc. & 0.604$\pm$0.259 & $>0.99$ (0.06) & 0.594$\pm$0.270 & 0.777 (0.11) & 31.95$\pm$56.83 & 0.009 (0.30) & 0.657 & 0.591 & 0.562 & 0.731 \\
 & VoxTell no text, full FT & 0.567$\pm$0.274 & 0.051 (0.24) & 0.553$\pm$0.286 & 0.018 (0.30) & 36.42$\pm$53.03 & $<0.001$ (0.39) & 0.564 & 0.613 & 0.525 & 0.678 \\
 & VoxTell text & \textbf{0.615$\pm$0.264} & - & 0.614$\pm$0.276 & - & 18.39$\pm$27.83 & - & \textbf{0.764} & 0.469 & \textbf{0.611} & \textbf{0.806} \\
\bottomrule
\end{tabular}

%% file: tables/external_cohort_results.tex
\begin{tabular}{lrlrlrlrrrr}
\toprule
Method & DSC & \(p_{\mathrm{adj}}\) (\(d_z\)) & NSD & \(p_{\mathrm{adj}}\) (\(d_z\)) & HD95 (mm) & \(p_{\mathrm{adj}}\) (\(d_z\)) & ET & ED & NCR & WT \\
\midrule
\multicolumn{11}{l}{\textbf{MEN (n=25)}} \\
nnU-Net & 0.678$\pm$0.207 & 0.544 (0.24) & 0.691$\pm$0.221 & 0.763 (0.18) & 11.10$\pm$14.74 & $>0.99$ (-0.09) & 0.884 & 0.595 & 0.556 & 0.852 \\
SAT-Pro & 0.723$\pm$0.187 & 0.833 (0.04) & \textbf{0.761$\pm$0.197} & 0.763 (-0.13) & \textbf{3.75$\pm$3.49} & 0.036 (-0.61) & \textbf{0.952} & \textbf{0.731} & 0.487 & \textbf{0.938} \\
VoxTell no text & 0.689$\pm$0.170 & 0.544 (0.28) & 0.679$\pm$0.191 & 0.266 (0.36) & 12.61$\pm$15.24 & $>0.99$ (0.18) & 0.888 & 0.595 & 0.584 & 0.769 \\
Pretrained T1c & 0.616$\pm$0.149 & $<0.001$ (0.98) & 0.596$\pm$0.144 & $<0.001$ (1.02) & 15.60$\pm$19.16 & 0.586 (0.30) & 0.930 & 0.351 & 0.566 & 0.770 \\
Blank T1c & 0.585$\pm$0.154 & $<0.001$ (1.18) & 0.570$\pm$0.144 & $<0.001$ (1.23) & 15.71$\pm$19.70 & 0.132 (0.49) & 0.936 & 0.337 & 0.484 & 0.801 \\
Contradictory T1c & 0.542$\pm$0.153 & $<0.001$ (1.11) & 0.525$\pm$0.149 & $<0.001$ (1.11) & 17.99$\pm$22.87 & 0.221 (0.42) & 0.925 & 0.334 & 0.367 & 0.730 \\
Correct T1c & 0.648$\pm$0.150 & 0.007 (0.71) & 0.637$\pm$0.146 & 0.001 (0.85) & 11.99$\pm$14.47 & $>0.99$ (0.08) & 0.934 & 0.359 & 0.653 & 0.768 \\
VoxTell text & \textbf{0.731$\pm$0.160} & - & 0.734$\pm$0.167 & - & 11.70$\pm$14.39 & - & 0.934 & 0.567 & \textbf{0.693} & 0.822 \\
\midrule
\multicolumn{11}{l}{\textbf{MET (n=25)}} \\
nnU-Net & 0.563$\pm$0.215 & 0.535 (0.23) & 0.574$\pm$0.232 & 0.520 (0.23) & 26.53$\pm$56.34 & 0.977 (-0.16) & 0.412 & 0.528 & 0.748 & 0.427 \\
SAT-Pro & 0.586$\pm$0.207 & 0.535 (0.13) & 0.629$\pm$0.223 & 0.948 (0.01) & \textbf{25.01$\pm$31.85} & 0.977 (-0.24) & 0.511 & \textbf{0.626} & 0.620 & 0.516 \\
VoxTell no text & 0.521$\pm$0.262 & 0.071 (0.48) & 0.533$\pm$0.271 & 0.044 (0.53) & 86.67$\pm$90.94 & 0.031 (0.63) & 0.560 & 0.384 & 0.620 & 0.463 \\
Pretrained T1c & 0.539$\pm$0.231 & 0.005 (0.75) & 0.538$\pm$0.233 & 0.001 (0.85) & 59.39$\pm$83.98 & 0.977 (0.23) & 0.603 & 0.336 & 0.678 & 0.486 \\
Blank T1c & 0.503$\pm$0.215 & 0.003 (0.82) & 0.507$\pm$0.241 & $<0.001$ (0.93) & 57.61$\pm$81.34 & 0.977 (0.22) & 0.594 & 0.323 & 0.593 & 0.517 \\
Contradictory T1c & 0.477$\pm$0.243 & 0.001 (0.88) & 0.470$\pm$0.255 & $<0.001$ (1.12) & 66.27$\pm$91.28 & 0.450 (0.39) & 0.608 & 0.307 & 0.517 & 0.436 \\
Correct T1c & 0.529$\pm$0.227 & 0.007 (0.70) & 0.537$\pm$0.235 & 0.001 (0.84) & 66.70$\pm$96.04 & 0.681 (0.31) & \textbf{0.612} & 0.339 & 0.636 & 0.507 \\
VoxTell text & \textbf{0.614$\pm$0.190} & - & \textbf{0.632$\pm$0.198} & - & 37.75$\pm$43.49 & - & \textbf{0.612} & 0.425 & \textbf{0.806} & \textbf{0.584} \\
\midrule
\multicolumn{11}{l}{\textbf{PEDs (n=25)}} \\
nnU-Net & 0.115$\pm$0.137 & 0.006 (0.76) & 0.121$\pm$0.142 & 0.102 (0.53) & 117.16$\pm$67.52 & $<0.001$ (1.41) & 0.041 & 0.270 & 0.034 & 0.281 \\
SAT-Pro & 0.322$\pm$0.160 & 0.331 (-0.20) & 0.318$\pm$0.163 & 0.816 (-0.27) & \textbf{17.58$\pm$10.84} & $>0.99$ (-0.12) & 0.110 & 0.517 & \textbf{0.339} & 0.782 \\
VoxTell no text & \textbf{0.366$\pm$0.167} & 0.199 (-0.34) & \textbf{0.332$\pm$0.164} & 0.816 (-0.29) & 23.91$\pm$21.12 & $>0.99$ (0.19) & 0.308 & \textbf{0.535} & 0.253 & 0.773 \\
Pretrained T1c & 0.175$\pm$0.160 & 0.008 (0.72) & 0.250$\pm$0.113 & 0.816 (0.06) & 23.05$\pm$22.75 & 0.160 (0.46) & 0.495 & 0.005 & 0.024 & 0.865 \\
Blank T1c & 0.224$\pm$0.158 & 0.159 (0.41) & 0.229$\pm$0.163 & 0.816 (0.21) & 22.21$\pm$22.11 & 0.612 (0.30) & 0.491 & 0.002 & 0.179 & 0.760 \\
Contradictory T1c & 0.216$\pm$0.164 & 0.045 (0.57) & 0.226$\pm$0.165 & 0.816 (0.26) & 23.69$\pm$22.84 & 0.062 (0.56) & 0.491 & 0.026 & 0.131 & 0.818 \\
Correct T1c & 0.231$\pm$0.188 & 0.045 (0.57) & 0.290$\pm$0.137 & 0.647 (-0.33) & 19.52$\pm$21.43 & $>0.99$ (-0.10) & 0.568 & 0.029 & 0.096 & 0.695 \\
VoxTell text & 0.280$\pm$0.150 & - & 0.259$\pm$0.170 & - & 20.00$\pm$20.97 & - & \textbf{0.635} & 0.029 & 0.176 & \textbf{0.900} \\
\midrule
\multicolumn{11}{l}{\textbf{UPENN-GBM (n=25)}} \\
nnU-Net & \textbf{0.854$\pm$0.162} & 0.002 (-0.74) & \textbf{0.863$\pm$0.156} & 0.004 (-0.73) & 4.08$\pm$5.04 & 0.975 (-0.01) & \textbf{0.886} & \textbf{0.891} & 0.784 & \textbf{0.936} \\
SAT-Pro & 0.810$\pm$0.146 & 0.002 (0.79) & 0.834$\pm$0.151 & $>0.99$ (-0.10) & \textbf{3.44$\pm$2.30} & 0.165 (-0.47) & 0.845 & 0.818 & 0.766 & 0.910 \\
VoxTell no text & 0.838$\pm$0.144 & 0.424 (-0.16) & 0.832$\pm$0.150 & $>0.99$ (-0.05) & 4.62$\pm$4.23 & 0.755 (0.20) & 0.873 & 0.848 & \textbf{0.793} & 0.918 \\
Pretrained T1c & 0.779$\pm$0.187 & $<0.001$ (1.16) & 0.744$\pm$0.185 & $<0.001$ (1.88) & 15.65$\pm$43.81 & 0.755 (0.27) & 0.871 & 0.721 & 0.746 & 0.808 \\
Blank T1c & 0.766$\pm$0.186 & $<0.001$ (1.37) & 0.733$\pm$0.186 & $<0.001$ (1.88) & 7.90$\pm$10.03 & 0.165 (0.46) & 0.860 & 0.687 & 0.751 & 0.796 \\
Contradictory T1c & 0.778$\pm$0.188 & $<0.001$ (1.18) & 0.738$\pm$0.189 & $<0.001$ (1.89) & 15.96$\pm$44.29 & 0.755 (0.27) & 0.873 & 0.710 & 0.752 & 0.799 \\
Correct T1c & 0.791$\pm$0.180 & $<0.001$ (1.17) & 0.765$\pm$0.180 & $<0.001$ (1.95) & 5.69$\pm$4.66 & 0.067 (0.56) & 0.876 & 0.736 & 0.762 & 0.821 \\
VoxTell text & 0.832$\pm$0.154 & - & 0.830$\pm$0.153 & - & 4.10$\pm$2.13 & - & 0.876 & 0.853 & 0.768 & 0.920 \\
\bottomrule
\end{tabular}

%% file: references.bib
@article{isensee2021nnunet,
  title={nnU-Net: a self-configuring method for deep learning-based biomedical image segmentation},
  author={Isensee, Fabian and Jaeger, Paul F. and Kohl, Simon A. A. and Petersen, Jens and Maier-Hein, Klaus H.},
  journal={Nature Methods},
  volume={18},
  pages={203--211},
  year={2021},
  doi={10.1038/s41592-020-01008-z}
}

@article{bakas2022upenn,
  title={The University of Pennsylvania glioblastoma (UPenn-GBM) cohort: advanced MRI, clinical, genomics, and radiomics},
  author={Bakas, Spyridon and Sako, Chiharu and Akbari, Hamed and Bilello, Michel and Sotiras, Aristeidis and Shukla, Gaurav and Rudie, Jeffrey D and Santamaria, Nelson F and Kazerooni, Anahita Fathi and Pati, Sarthak and others},
  journal={Scientific Data},
  volume={9},
  number={1},
  pages={453},
  year={2022},
  publisher={Nature Publishing Group}
}

@article{zhao2025sat,
  title={Large-vocabulary segmentation for medical images with text prompts},
  author={Zhao, Ziheng and Zhang, Yao and Wu, Chaoyi and Zhang, Xiaoman and Zhou, Xiao and Zhang, Ya and Wang, Yanfeng and Xie, Weidi},
  journal={npj Digital Medicine},
  volume={8},
  number={566},
  year={2025},
  doi={10.1038/s41746-025-01964-w}
}

@misc{rokuss2025voxtell,
  title={VoxTell: Free-Text Promptable Universal 3D Medical Image Segmentation},
  author={Rokuss, Maximilian and Langenberg, Moritz and Kirchhoff, Yannick and Isensee, Fabian and Hamm, Benjamin and Ulrich, Constantin and Regnery, Sebastian and Bauer, Lukas and Katsigiannopulos, Efthimios and Norajitra, Tobias and Maier-Hein, Klaus},
  year={2025},
  eprint={2511.11450},
  archivePrefix={arXiv},
  primaryClass={cs.CV},
  doi={10.48550/arXiv.2511.11450}
}

@article{menze2015brats,
  title={The Multimodal Brain Tumor Image Segmentation Benchmark (BRATS)},
  author={Menze, Bjoern H. and Jakab, Andras and Bauer, Stefan and Kalpathy-Cramer, Jayashree and Farahani, Keyvan and Kirby, Justin and Burren, Yuliya and Porz, Nicole and Slotboom, Johannes and Wiest, Roland and others},
  journal={IEEE Transactions on Medical Imaging},
  volume={34},
  number={10},
  pages={1993--2024},
  year={2015},
  doi={10.1109/TMI.2014.2377694}
}

@article{bakas2018brats,
  title={Identifying the best machine learning algorithms for brain tumor segmentation, progression assessment, and overall survival prediction in the BRATS challenge},
  author={Bakas, Spyridon and Akbari, Hamed and Sotiras, Aristeidis and Bilello, Michel and Rozycki, Martin and Kirby, Justin S. and Freymann, John B. and Farahani, Keyvan and Davatzikos, Christos},
  journal={arXiv preprint arXiv:1811.02629},
  year={2018}
}

@article{pereira2016bratscnn,
  title={Brain Tumor Segmentation Using Convolutional Neural Networks in MRI Images},
  author={Pereira, Sergio and Pinto, Adriano and Alves, Victor and Silva, Carlos A.},
  journal={IEEE Transactions on Medical Imaging},
  volume={35},
  number={5},
  pages={1240--1251},
  year={2016},
  doi={10.1109/TMI.2016.2538465}
}

@article{havaei2017brain,
  title={Brain tumor segmentation with Deep Neural Networks},
  author={Havaei, Mohammad and Davy, Axel and Warde-Farley, David and Biard, Antoine and Courville, Aaron and Bengio, Yoshua and Pal, Chris and Jodoin, Pierre-Marc and Larochelle, Hugo},
  journal={Medical Image Analysis},
  volume={35},
  pages={18--31},
  year={2017},
  doi={10.1016/j.media.2016.05.004}
}

@article{kamnitsas2017efficient,
  title={Efficient multi-scale 3D CNN with fully connected CRF for accurate brain lesion segmentation},
  author={Kamnitsas, Konstantinos and Ledig, Christian and Newcombe, Virginia F. J. and Simpson, Joanna P. and Kane, Andrew D. and Menon, David K. and Rueckert, Daniel and Glocker, Ben},
  journal={Medical Image Analysis},
  volume={36},
  pages={61--78},
  year={2017},
  doi={10.1016/j.media.2016.10.004}
}

@article{myronenko20183d,
  title={3D MRI brain tumor segmentation using autoencoder regularization},
  author={Myronenko, Andriy},
  journal={Brainlesion: Glioma, Multiple Sclerosis, Stroke and Traumatic Brain Injuries},
  pages={311--320},
  year={2018}
}

@article{hatamizadeh2022unetr,
  title={UNETR: Transformers for 3D Medical Image Segmentation},
  author={Hatamizadeh, Ali and Yang, Dong and Roth, Holger and Xu, Daguang},
  journal={Proceedings of the IEEE/CVF Winter Conference on Applications of Computer Vision},
  pages={574--584},
  year={2022}
}

@article{tang2022swinunetr,
  title={Self-supervised pre-training of Swin Transformers for 3D medical image analysis},
  author={Tang, Yucheng and Yang, Dong and Li, Wenqi and Roth, Holger R. and Landman, Bennett and Xu, Daguang and Nath, Vishwesh and Hatamizadeh, Ali},
  journal={Proceedings of the IEEE/CVF Conference on Computer Vision and Pattern Recognition},
  pages={20730--20740},
  year={2022}
}

@article{liu2023survey,
  title={Deep learning based brain tumor segmentation: a survey},
  author={Liu, Zhihua and Tong, Lei and Chen, Long and Jiang, Zheheng and Zhou, Feixiang and Zhang, Qianni and Zhang, Xiangrong and Jin, Yaochu and Zhou, Huiyu},
  journal={Complex \& Intelligent Systems},
  volume={9},
  pages={1001--1026},
  year={2023},
  doi={10.1007/s40747-022-00815-5}
}

@misc{wang2025triad,
title = {Vision foundation model for 3D magnetic resonance imaging segmentation, classification, and registration},
journal = {Medical Image Analysis},
volume = {110},
pages = {103992},
year = {2026},
issn = {1361-8415},
doi = {https://doi.org/10.1016/j.media.2026.103992},
author = {Shansong Wang and Mojtaba Safari and Qiang Li and Chih-Wei Chang and Richard {LJ Qiu} and Justin Roper and David S. Yu and Xiaofeng Yang},
}

@misc{he2024vista3d,
  title={VISTA3D: A Unified Segmentation Foundation Model For 3D Medical Imaging},
  author={He, Yufan and Guo, Pengfei and Tang, Yucheng and Myronenko, Andriy and Nath, Vishwesh and Xu, Ziyue and Yang, Dong and Zhao, Can and Simon, Benjamin and Belue, Mason and Harmon, Stephanie and Turkbey, Baris and Xu, Daguang and Li, Wenqi},
  year={2024},
  eprint={2406.05285},
  archivePrefix={arXiv},
  primaryClass={eess.IV}
}

@misc{li2025meddinov3,
  title={MedDINOv3: How to Adapt Vision Foundation Models for Medical Image Segmentation?},
  author={Li, Yuheng and Wu, Yizhou and Lai, Yuxiang and Hu, Mingzhe and Yang, Xiaofeng},
  year={2025},
  eprint={2509.02379},
  archivePrefix={arXiv},
  primaryClass={cs.CV}
}
